\documentclass{article}
\usepackage{iclr2027_conference,times}
\iclrfinalcopy

\usepackage{hyperref,url,booktabs,multirow,array,graphicx}

\usepackage{amsmath,amsfonts,bm}

\def\eqref#1{equation~\ref{#1}}

\def\1{\bm{1}}

\DeclareMathAlphabet{\mathsfit}{\encodingdefault}{\sfdefault}{m}{sl}
\SetMathAlphabet{\mathsfit}{bold}{\encodingdefault}{\sfdefault}{bx}{n}

\definecolor{cvprblue}{rgb}{0.21,0.49,0.74}
  
\newcommand{\ourmethod}{Switch-SWE}

\usepackage{amssymb}
\usepackage[table]{xcolor}   
\usepackage{pifont}          
\usepackage{wasysym}         
\newcommand{\cmark}{\ding{51}}
\newcommand{\xmark}{\ding{55}}
\newcommand{\pmark}{\raisebox{0.02ex}{\scalebox{0.88}{\LEFTcircle}}}

\title{Switch-SWE: Do Coding Agents Hold Focus When Switching Between Unfinished Tasks?}

\author{
\renewcommand{\arraystretch}{1.05}
\begin{tabular}{c}
Shuochen Chang\textsuperscript{1,*} \quad
Qianli Ma\textsuperscript{1,*} \quad
Yibo Miao\textsuperscript{1} \quad
Qingyang Liu\textsuperscript{1} \quad
Bingjie Gao\textsuperscript{1} \\[5pt]
Haonan Zhao\textsuperscript{1} \quad
Siyu Sun\textsuperscript{1} \quad
Yulin Sun\textsuperscript{1} \quad
Han Zhao\textsuperscript{1} \quad
Jiangtong Li\textsuperscript{2,\dag} \\[7pt]
\mdseries
\textsuperscript{1}Shanghai Jiao Tong University
\qquad
\textsuperscript{2}Tongji University \\[3pt]
\mdseries
\textsuperscript{*}\textit{Equal contribution.}
\qquad
\textsuperscript{\dag}\textit{Corresponding author.} \\[4pt]
\mdseries\small
\texttt{\{csc1332741686,mqlqianli\}@sjtu.edu.cn}
\quad
\texttt{jiangtongli@tongji.edu.cn}
\end{tabular}
}

\begin{document}

\maketitle
\lhead{Preprint.}

\begin{abstract}
Software development sessions often involve several unfinished tasks rather than one issue carried through to completion.
As coding agents take on repository-level development, they need to recover an earlier task's requirements and partial progress after intervening work.
Existing repository-level benchmarks typically score one issue-to-patch episode and do not directly measure whether an agent can resume a task after advancing other unfinished work in the same repository.
We introduce \textbf{\ourmethod}, a benchmark for focus-task resumption in sessions that interleave unfinished tasks drawn from the same source repository, with a persistent checkout for each task.
Each focus task is evaluated in matched uninterrupted and switched sessions with identical focus requirements and executable verifiers.
The switched session interleaves ordered stages from other unfinished tasks in the same repository, while only the final focus patch is scored.
\ourmethod~contains 200 focus tasks from 19 Python repositories, reconstructed from issue, pull-request, and review histories and rendered as 400 matched session specifications.
Across 7 coding agents, interleaved unfinished work produces an average focus-task resolution gap of 11.5 percentage points, ranging from 6.0 to 15.0 percentage points.
Trajectory analysis identifies cross-task requirement confusion and loss of task state as recurrent resumption behaviors.
We introduce a plug-in context-management module that selects task-relevant history and maintains concise task-state summaries.
On OpenCode, the module improves focus-task completion under same-repository task switching by 5.5 to 15.5 percentage points across the evaluated models. Code is available at \url{github.com/Miraclecsc/Switch-SWE}.
\end{abstract}
\section{Introduction}
\label{sec:intro}

Coding agents are expanding the scope of automated software development, combining code generation with repository exploration, code editing, and test execution to address increasingly complex tasks~\citep{yang2024sweagent,zhang2024autocoderover,wang2025openhands,xia2024agentless,chang2026struggle}.
As agents take on broader development work, users can maintain several requests within one continuous session.
A feature request may remain unfinished while the agent handles a separate bug or review request before returning to the earlier work.
Reliable collaboration then requires the agent to recover the earlier task's requirements and partial progress while keeping them separate from intervening work.
Recent advances in coding benchmarks make this capability amenable to systematic evaluation~\citep{chen2021codex,chang2026unlocking,jimenez2024swebench,zan2025multiswebench,yang2024swebenchmm,liu2024repobench}.

Existing benchmarks have expanded coding-agent evaluation in complementary directions.
SWE-bench evaluates repository-level issue resolution~\citep{jimenez2024swebench}.
SWE-Touch evaluates adaptation to user code edits within a coding task~\citep{tan2026swetouch}.
SWE-ContextBench evaluates the reuse of context from related tasks~\citep{zhu2026swecontextbench}.
InterruptBench examines adaptation to changing user goals in web navigation~\cite{zou2026interruptbench}.
Their settings do not directly isolate whether a coding agent can recover and complete an unfinished repository-level task after advancing other unfinished tasks in the same continuous session.
This raises a central question: \textit{can a coding agent reliably resume and complete an unfinished task after working on other unfinished tasks?}

To answer this question, we introduce \textbf{\ourmethod}, a benchmark for session-level resumption after an agent advances other unfinished tasks from the same source repository.
Each focus task is evaluated in matched sessions.
The uninterrupted control presents the focus task's substantive stages in order.
The switched session inserts ordered stages from other unfinished tasks in that repository before the focus task is submitted.
Each task retains its own persistent checkout, preserving its partial work while keeping the comparison focused on session-level task recovery rather than on-disk conflicts.
The focus requirements and executable verifier are identical across the matched pair, and only the final focus patch is scored.
This paired design measures how interleaved unfinished work affects an agent's ability to resume and complete the focus task.

\begin{figure}[t!]
            \centering
            \includegraphics[width=\linewidth]{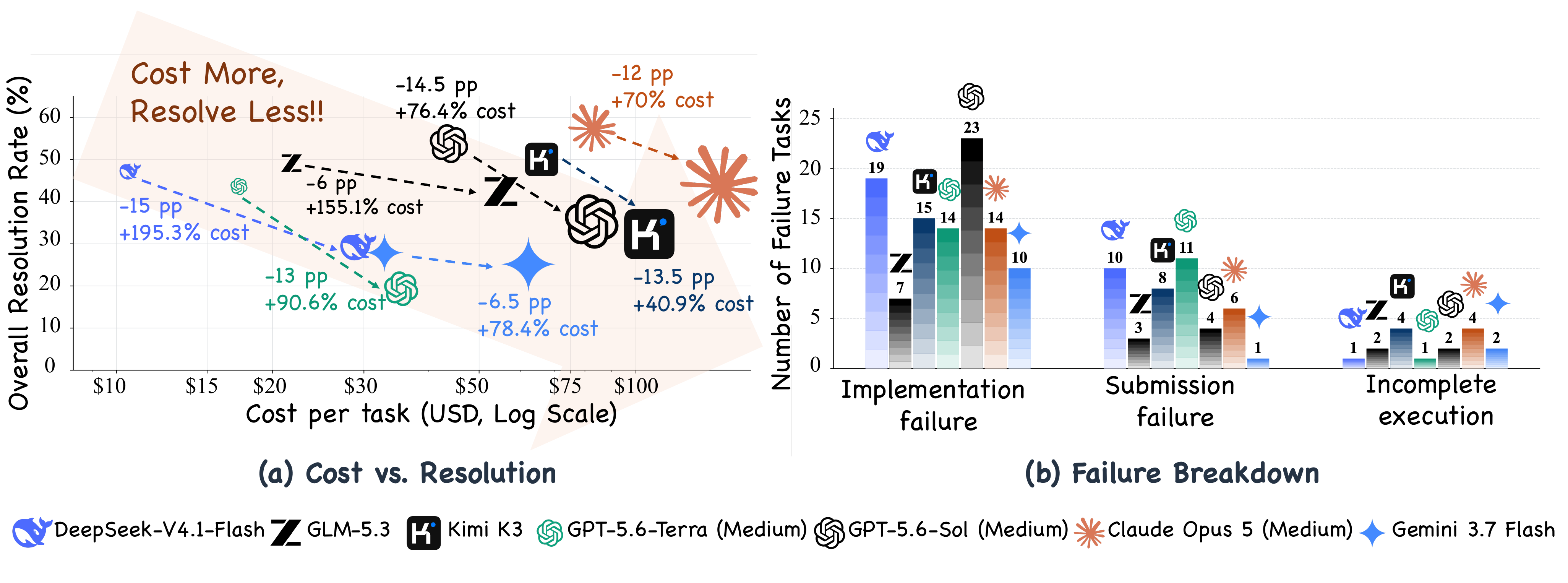}
            \caption{
            \textbf{Task switching increases cost while reducing completion.}
(a) Moving from uninterrupted to switched execution lowers focus-task resolution and raises cost across all seven configurations.
(b) Failures span implementation, submission, and incomplete execution.}
     \label{fig:intro}
\end{figure}

\ourmethod~contains 200 focus tasks from 19 Python repositories, each forming a matched uninterrupted control and switched session, for 400 session specifications.
Each focus task contains substantive stages reconstructed from real issue, pull-request, and review histories.
Every switched session interleaves the focus task with four unfinished tasks from the same repository.
We assign interruptions to the same source subsystem as the focus task or a different subsystem, enabling analysis of how source locality relates to task resumption.
Alongside executable focus-task verification, \ourmethod~records recovery costs and cross-task errors to characterize task-resumption behavior.

We evaluate 7 coding agents on \ourmethod.
Across these agents, interleaved unfinished work yields an average focus-task resolution gap of 11.5 percentage points relative to matched uninterrupted sessions.
The gap ranges from 6.0 to 15.0 percentage points across agents, revealing a substantial task-resumption challenge despite identical focus requirements, executable verifiers, and task-specific workspaces in every matched pair.
Figure~\ref{fig:intro}(a) shows that lower resolution accompanies higher execution cost, while Figure~\ref{fig:intro}(b) breaks down failures into implementation, submission, and incomplete execution.
Trajectory analysis identifies cross-task requirement confusion and loss of task state as recurrent resumption behaviors.

These trace-level patterns point to two context-recovery needs: isolating the history relevant to the active task and recovering its accumulated state.
We introduce a plug-in context-management module that addresses these needs through task-relevant history selection and concise task-state summaries.
The module leaves the agent's native execution workflow and persistent per-task checkouts unchanged.
On OpenCode, the module improves focus-task completion under same-repository task switching by 5.5 to 15.5 percentage points across the evaluated models.
Our contributions are:
\begin{itemize}
\item \textbf{Setting.} We formulate same-repository task resumption as an evaluation setting with matched uninterrupted and interleaved sessions.
\item \textbf{Benchmark.} We construct \ourmethod~from executable multi-stage repair tasks and real issue, pull-request, and review histories.
\item \textbf{Evaluation.} We provide paired measures of focus-task resolution, source locality, and resumption behavior.
\item \textbf{Method.} We introduce a plug-in context manager for task-relevant history and task-state recovery, and show improved task completion on OpenCode.
\end{itemize}

\section{Related Work}
\label{sec:related_work}

\noindent\textbf{Repository-Level Coding Benchmarks.}~
Repository-level coding benchmarks evaluate how models use project context to complete code, resolve issues, and perform broader software engineering tasks~\citep{liu2024repobench,jimenez2024swebench,rashid2025swepolybench,miserendino2025swelancer,stepfun2026inside}.
SWE-bench established executable evaluation of repository-level issue resolution, scoring generated patches against real GitHub pull-request tests~\citep{jimenez2024swebench}.
SWE-bench-Live continuously curates recent issues~\citep{zhang2025swebenchlive}.
Multi-SWE-bench extends evaluation across programming languages~\citep{zan2025multiswebench,rashid2025swepolybench}.
DeepSWE constructs original engineering tasks with purpose-built behavioral verifiers~\citep{huang2026deepswe}.
These benchmarks establish executable repository-level issue resolution as our foundation.
Switch-SWE changes the evaluation unit from one issue-to-patch episode to a same-repository session where agents repeatedly suspend and resume unfinished tasks.

\noindent\textbf{Long-Horizon and Continuous Software Development.}~
Recent benchmarks evaluate long-horizon or continuous software development~\citep{zhao2025commit0,merrill2026terminalbench,desai2026swemarathon,deng2026swemilestone}.
Terminal-Bench evaluates agents on tasks in command-line environments~\citep{merrill2026terminalbench}, while SWE-Marathon targets project-scale work with extended execution horizons and multi-layer verification~\citep{desai2026swemarathon}.
SWE-EVO constructs multi-step software evolution tasks from release histories~\citep{le2026sweevo}.
SWE-Milestone evaluates sequences of dependent milestones in a persistent codebase~\citep{deng2026swemilestone}, and ChainSWE studies sequential bug fixing with accumulated code changes and conversational history~\citep{jin2026chainswe}.
These settings emphasize sustained execution, dependent milestones, or cumulative changes across development.
Switch-SWE evaluates the same agent as it pauses a still-unfinished focus task, advances other unfinished tasks within the same repository, and returns to it.

\noindent\textbf{Task Recovery and Interactive Coding.}~
Prior work studies how developers and agents recover task context after interruptions and adapt their work as requirements and user feedback unfold~\citep{parnin2011resumption,parnin2010evaluating,laban2025multiturnlost,kersten2006using,chang2026d3tom}.
Task recovery is studied directly by ContinuityBench, which resumes interrupted agents from preserved environment states under different handoff conditions~\citep{gulati2026continuitybench}, and Handoff Debt, which measures the cost of taking over another agent's unfinished coding task~\citep{kc2026handoffdebt}.
These settings study recovery from preserved states or transfer of unfinished work between agents.
Interactive benchmarks address a different form of adaptation as a task unfolds.
SWE-Together reconstructs real user-agent coding sessions with a reactive user simulator~\citep{wu2026swetogether}, while SWE-INTERACT progressively reveals requirements and provides user feedback~\citep{raghavendra2026sweinteract}.
SWE-Touch introduces user edits to the agent's active coding workspace~\citep{tan2026swetouch}.
Beyond coding, InterruptBench studies interruptions that add, revise, or retract goals during web navigation~\citep{zou2026interruptbench}.
Switch-SWE keeps one agent in a continuous session, where intervening work consists of other unfinished coding tasks in the same repository rather than changes to the active task or a handoff to a successor.

\noindent\textbf{Context Management for Task Resumption.}~
Context-management methods organize interaction histories and maintain explicit memories for reliable behavior over extended sessions~\citep{packer2023memgpt,xu2026mem,chhikara2025mem0,lindenbauer2025complexity,liu2026context,zhang2026agentic,liu2024lostmiddle}.
SWE-ContextBench evaluates coding agents' retrieval and reuse of experience from previously executed, related tasks~\citep{zhu2026swecontextbench}.
In multi-task computer-use settings, CORPGEN studies interleaved workloads and uses task isolation and structured memory to manage competing contexts~\citep{jaye2026corpgen}.
These works motivate context management, but Switch-SWE exposes a complementary setting where an agent must recover unfinished work from a shared session within the same repository.
Our plug-in context-management module retrieves task-relevant history and maintains task-specific progress summaries to support resumption.
\section{\ourmethod}
\label{sec:benchmark}

\begin{figure}[t]
    \centering
    \includegraphics[width=\linewidth]{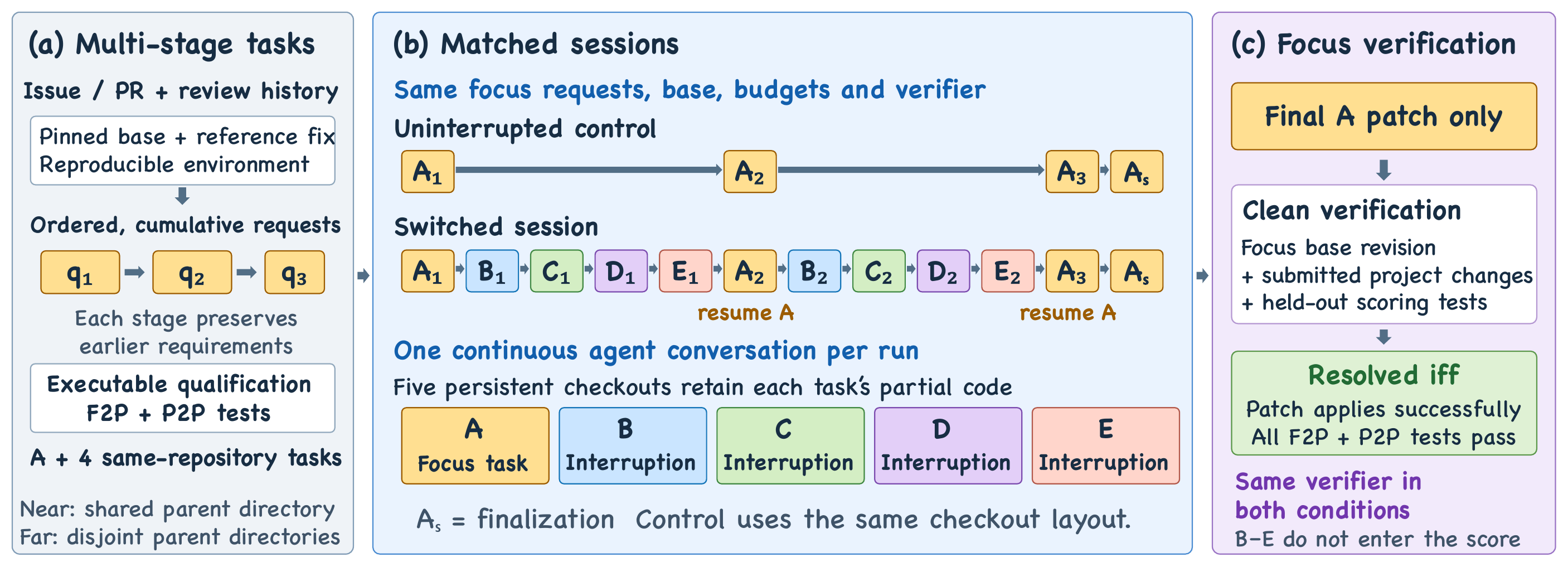}
    \caption{
    \textbf{Overview of \ourmethod.}
    \textbf{(a)} Repository-level tasks are represented as ordered,
    cumulative requests grounded in issue, pull-request, and review
    histories.
    \textbf{(b)} Each focus task is evaluated in matched uninterrupted
    and switched sessions. The switched session interleaves unfinished
    same-repository tasks as interruption.
    \textbf{(c)} Only the final focus-task patch is scored using the same
    executable verifier in both conditions. Interruption-task
    implementations do not enter primary score.
    }
    \label{fig:construction}
\end{figure}

\subsection{Matched Session Design}
\label{sec:session_design}

\ourmethod~evaluates focus-task resumption by interleaving the focus task with four unfinished same-repository tasks in a shared session, each with a persistent checkout.
We define matched schedules preserving the focus request sequence and verifier, describe task construction and locality-based interruption selection, and present final focus-patch verification and benchmark statistics. Figure~\ref{fig:construction} provides an overview of this design.
\\\textbf{Multi-stage tasks.}
We represent each task as $t=(R_t,Q_t,V_t)$, where $R_t$ is its base repository snapshot, $Q_t=(q_{t,1},\ldots,q_{t,k_t})$ is its ordered sequence of substantive requests, and $V_t$ is its executable verifier.
Each focus task contains substantive requests, allowing the schedule to suspend and resume it multiple times.
A requirement-free finalization request asks the agent to complete the implementation and is excluded from the substantive-stage count.
\\\textbf{Paired schedules.}
For a focus task $A$, the uninterrupted control delivers its substantive requests consecutively, followed by finalization:
\begin{equation}
\mathcal{S}_{\mathrm{base}}(A)
=
(A_1,\ldots,A_{k_A},A_{\mathrm{submit}}).
\label{eq:baseline_schedule}
\end{equation}
The switched schedule interleaves ordered work on four other tasks, $B$, $C$, $D$, and $E$, between requests for $A$.
For a three-stage focus task, one illustrative schedule is
\begin{equation}
\begin{aligned}
\mathcal{S}_{\mathrm{switch}}(A)=(
A_1,B_1,C_1,D_1,E_1,A_2,
B_2,C_2,D_2,E_2,A_3,A_{\mathrm{submit}}).
\end{aligned}
\label{eq:switched_schedule}
\end{equation}
Each interruption task contributes an ordered request prefix delivered in one or more chunks.
Interruption work is interleaved with focus requests, and focus finalization is always the last request.
Schedules are fixed before evaluation and preserve each task's request order; 
detailed schedule-generation rules are provided in the Appendix~\ref{app:schedule_generation}.

Let $\pi_A$ project a schedule onto the requests for task $A$.
The matched pair satisfies
\begin{equation}
\pi_A\bigl(\mathcal{S}_{\mathrm{switch}}(A)\bigr)
=
\mathcal{S}_{\mathrm{base}}(A).
\label{eq:matched_projection}
\end{equation}
This equality holds for all focus-message payloads and their order, including finalization.
The focus snapshot, verifier, checkout layout, and matched focus-stage budgets are fixed across conditions.
The paired outcome measures focus-task resolution under the specified interleaved-work schedule.
\\\textbf{Persistent workspaces, shared conversation.}
Each task has a persistent checkout at its own base revision.
Both conditions initialize the same five role-neutral checkout paths; in the uninterrupted control, only focus-task requests are delivered.
Each checkout retains its task's code across stages, while the same native agent conversation accumulates requests, actions, and observations within each run.
The uninterrupted and switched runs start independently with clean files, conversation state, and agent memory.
Each evaluated agent follows its native context-compaction policy.
\\\textbf{Request delivery.}
The controller delivers one request at a time with the repository name, active checkout path, and cleaned requirement text.
The path and request identify the active task; future requests, private focus/interruption labels, task-specific recovery summaries, and verifier feedback are withheld.
Request presentation is detailed in Appendix~\ref{app:requirement_reconstruction}; and stage-ending rules and execution budgets are specified in Appendix~\ref{sec:experimental_setup}.

\subsection{Constructing Executable Multi-Stage Tasks}
\label{sec:construction}

Each task combines a reproducible repository repair, a source-grounded sequence of requirements reconstructed from its development history, and a requirement-aligned executable verifier.
We construct multi-stage task specifications from issue, pull-request, and review material, combining source-grounded requirements with behavioral extensions and executable tests where needed.
\\\textbf{Executable repair instances.}
We collect candidates from the SWE-bench family and additional merged pull requests in Python repositories~\citep{jimenez2024swebench,zhang2025swebenchlive}.
Each candidate must have a reproducible base revision, traceable issue or pull-request records, a reference fix, and a pinned executable environment.
We separate each reference change into implementation and test patches, then run designated tests on the base revision with its test patch and on that same state after applying the reference fix.
We retain tasks only when outcomes are stable, with at least one fail-to-pass test (fails on the base and passes with the reference fix) and designated pass-to-pass tests (pass in both states) showing no regression.
Detailed test-classification rules and execution commands are provided in Appendix~\ref{app:test_classification}.
\\\textbf{Requirements from development histories.}
The first Stage is drawn from the linked issue title and body, or from the pull-request description when no issue is available.
Later stages follow substantive pre-merge review rounds chronologically, consolidating comments that express one coherent requirement update.
Automated messages, duplicate comments, approvals without new requirements, and progress-only reports are excluded.
We preserve requested behavior and relevant public interfaces while removing reference-code excerpts and instructions that rely on unavailable intermediate patches.
A stage is retained only if its essential meaning remains interpretable from earlier requests and the agent-visible repository.
Reconstruction and review criteria are in Appendix~\ref{app:requirement_reconstruction}.
\\\textbf{Requirement-verifier alignment.}
We audit the request sequence against the scoring tests, linking each test's expected outcome to a requirement or public behavior stated in the specification and checking that every stage expresses a substantive request understandable without the reference implementation.
Cases requiring clarification are revised using their source records and re-validated; cases without a source-supported resolution are excluded.
Stage-to-source and requirement-to-test mappings are retained in the benchmark metadata.
\\\textbf{Provenance and evaluation assets.}
\label{sec:hygiene}
The benchmark manifest records source IDs, base revisions, cleaned requests, environment recipes, test IDs, and validation outcomes, linking each request to its source history and each score to a reproducible verifier.
Reference implementations, held-out test patches, and scoring assets remain outside agent-accessible works

\subsection{Assembling Same-Repository Interruptions}
\label{sec:locality}
Each session pairs a focus task with four distinct interruption tasks from the same repository; we compare resumption across source-directory locality groups.
\\\textbf{Source-directory locality.}
For task $t$, let $C_t$ be the repository-relative paths of source files modified by its reference fix.
We exclude test files, documentation, and other non-source artifacts using a repository-specific classification fixed before selection, and omit tasks with no changed source files.
Let $D_t=\{\operatorname{parent}(p):p\in C_t\}$ be the set of immediate parent directories of those files.
A focus-interruption pair $(f,b)$ is eligible only if $C_f\cap C_b=\varnothing$.
For eligible pairs, define locality as follows:
\begin{equation}
\operatorname{locality}(f,b)=
\begin{cases}
\mathrm{near}, & D_f\cap D_b\neq\varnothing,\\
\mathrm{far},  & D_f\cap D_b=\varnothing.
\end{cases}
\label{eq:locality_definition}
\end{equation}
Near pairs modify distinct files in at least one shared immediate directory, whereas far pairs have no shared immediate directory.
\\\textbf{Interruption selection.}
We select 100 near and 100 far focus tasks using task metadata and earlier execution outcomes for a fixed diagnostic cohort.
For each focus, we randomly sample four distinct same-repository interruption tasks satisfying the assigned locality and source-file disjointness criteria.
Focus-task IDs are unique across the benchmark; interruption-task reuse across sessions is recorded in the manifest.
We pair uninterrupted and switched runs by focus task, then compare their resolution gaps across locality groups alongside control performance and task composition.
\\\textbf{Schedule assembly.}
Given the case seed, the scheduler selects an ordered request prefix for each interruption and places its chunks among focus requests before finalization.
Automated checks verify stage order, focus-stage coverage, and finalization as the last request, and record interruption placement and resumption patterns.
Focus-message wording differences and schedule-validation results are documented in Appendix~\ref{app:schedule_generation}.
The released session specifications record complete schedules, task-to-workspace mappings, locality assignments, and seeds.
The selected cohort and session schedules are fixed before the reported evaluation.

\subsection{Executable Focus-Task Verification}
\label{sec:verification}
Both conditions are scored solely on whether the final focus implementation resolves its task under the same held-out tests and verification procedure.
\\\textbf{Capturing the final submission.}
At finalization, the evaluator parses the agent's submitted focus-task patch.
When used, packaging normalization extracts the structured submission from its outer presentation format without changing the decoded patch.
Submission handling is detailed in Appendix~\ref{sec:evaluation_protocol}.
The submitted patch may modify any eligible project file, not just in the reference fix.
\\\textbf{Held-out execution.}
The verifier applies the submitted non-test project changes and held-out scoring-test patch to the focus base revision in a clean environment, then runs the fixed command.
Agent-authored tests may support development but do not replace the held-out scoring tests.
Let $\mathcal{F}_f$ and $\mathcal{P}_f$ denote the validated fail-to-pass and pass-to-pass sets for focus task $f$.
Resolution requires successful artifact application and passage of every test in their union:
\begin{equation}
\operatorname{Resolved}(f)=
\mathbb{I}\!\left[
\operatorname{Apply}(f)=1
\;\land\;
\bigwedge_{u\in\mathcal{F}_f\cup\mathcal{P}_f}
\operatorname{Pass}(u)=1
\right].
\label{eq:resolution_criterion}
\end{equation}
Only tests that are collected, executed, and pass count toward resolution; artifact-application failures, candidate-induced test failures, and timeouts yield unresolved outcomes.
Confirmed infrastructure faults trigger re-evaluation of the same saved submission artifact under Appendix~\ref{sec:experimental_setup}.
Intermediate workspace states and interruption implementations are retained for behavioral analysis but excluded from the primary resolution score, and private intermediate test results are withheld from agents.
Scoring and aggregation procedures are detailed in Appendix~\ref{sec:evaluation_protocol}.

\subsection{Benchmark Profile}
\label{sec:dataset_characteristics}

\ourmethod~contains 200 focus tasks from 19 Python repositories, each yielding a matched uninterrupted control and switched session, for 400 session specifications.
Each switched session interleaves the focus task with four distinct unfinished same-repository tasks.
Table~\ref{tab:session_stats} summarizes task coverage and median switched-session workload.
A workspace switch occurs when consecutive requests designate different checkouts.
A focus return is a substantive focus request immediately following work in another checkout; finalization requests are excluded.
These metrics describe the scheduled workload, while the conversation history contains agent actions and tool observations.

\section{Experimental Results}
\label{sec:experiments}
\label{sec:results}

\textbf{Experimental Setup.}
We evaluate seven model-agent configurations across five harnesses, using the highest available reasoning effort unless otherwise specified in the model name.
All runs use Harbor with one continuous agent session per run.
Each agent invocation has a 900-second budget, with at most three timeout continuations across the entire session.
Each configuration is evaluated on the same 200 focus tasks under matched uninterrupted and switched conditions, including 100 near and 100 far tasks.
Following SWE-bench~\citep{jimenez2024swebench}, we score only the final focus-task patch: a task is resolved when the submitted patch passes all designated fail-to-pass and pass-to-pass tests.
Runs without a final submission or scoring result count as unresolved.
Agent configurations, execution settings, and the patch-evaluation protocol are detailed in Appendix~\ref{sec:experimental_setup}.

\subsection{Focus-Task Resolution}
\label{sec:main_results}

Table~\ref{tab:main_results} presents overall resolution and its source-locality breakdown.
We define the resolution gap as uninterrupted - switched resolution.
All configurations exhibit positive gaps, averaging 11.5\%.
\\\textbf{Task switching reveals a consistent resumption gap.}
Gaps range from 6.0 to 15.0 \% .
DeepSeek-V4.1-Flash with OpenCode drops from 46.5\% uninterrupted resolution to 31.5\% under switching, a relative reduction of 32.3\%.
GPT-5.6 Sol (medium) with Codex and Kimi K3 with KimiCode show gaps of 14.5 and 13.5 \% , respectively.
However, a small gap alone does not imply strong switched performance: Gemini 3.7 Flash has a gap of 6.5 \%  but the lowest switched resolution at 23.0\%.
The gap therefore needs to be read alongside absolute resolution.
\\\textbf{Task switching changes the performance ordering.}
Claude Opus 5 with Claude Code achieves the highest uninterrupted resolution at 53.5\%, whereas GLM-5.3 with OpenCode leads under switching at 43.0\%.
GLM-5.3 combines the smallest gap, at 6.0 \% , with the highest switched resolution.
Compared with Kimi K3, their uninterrupted rates differ by only 1.0 \%  (49.0\% versus 48.0\%), but their switched rates differ by 8.5 \%  (43.0\% versus 34.5\%).
Thus, configurations with similar uninterrupted performance can diverge after intervening work, showing why uninterrupted resolution is insufficient to characterize task-resumption performance.

\begin{table}[t]
\centering
\caption{\textbf{Focus-task resolution with and without task switching.}
Overall includes 200 tasks; Near and Far include 100 each.
Unint.\ and Switch report resolution (\%); Gap is Unint.\ minus Switch (\% ).
Within each group, bold marks the highest Switch and lowest Gap; underlining marks the lowest Switch and highest Gap, including ties.}
\label{tab:main_results}
\label{tab:locality_results}
\begingroup
\small
\setlength{\tabcolsep}{4pt}
\renewcommand{\arraystretch}{1.16}
\resizebox{\linewidth}{!}{%
\begin{tabular}{@{}llrrr@{\hspace{9pt}}rrr@{\hspace{9pt}}rrr@{}}
\toprule
\multirow{2}{*}{\textbf{Model}}
& \multirow{2}{*}{\textbf{Harness}}
& \multicolumn{3}{c}{\textbf{Overall}}
& \multicolumn{3}{c}{\textbf{Near}}
& \multicolumn{3}{c}{\textbf{Far}} \\
\cmidrule(lr){3-5}
\cmidrule(lr){6-8}
\cmidrule(lr){9-11}
& & Unint. & Switch$\uparrow$ & Gap$\downarrow$
  & Unint. & Switch$\uparrow$ & Gap$\downarrow$
  & Unint. & Switch$\uparrow$ & Gap$\downarrow$ \\
\midrule
DeepSeek-V4.1-Flash
& OpenCode
& 46.5 & 31.5 & \underline{15.0}
& 45.0 & 34.0 & 11.0
& 48.0 & 29.0 & \underline{19.0} \\
GLM-5.3
& OpenCode
& 49.0 & \textbf{43.0} & \textbf{6.0}
& 47.0 & \textbf{46.0} & \textbf{1.0}
& 51.0 & \textbf{40.0} & \textbf{11.0} \\
Kimi K3
& KimiCode
& 48.0 & 34.5 & 13.5
& 47.0 & 35.0 & 12.0
& 49.0 & 34.0 & 15.0 \\
GPT-5.6 Terra (medium)
& Codex
& 41.0 & 28.0 & 13.0
& 42.0 & 32.0 & 10.0
& 40.0 & 24.0 & 16.0 \\
GPT-5.6 Sol (medium)
& Codex
& 52.5 & 38.0 & 14.5
& 53.0 & 40.0 & \underline{13.0}
& 52.0 & 36.0 & 16.0 \\
Claude Opus 5 (medium)
& Claude Code
& 53.5 & 41.5 & 12.0
& 52.0 & 44.0 & 8.0
& 55.0 & 39.0 & 16.0 \\
Gemini 3.7 Flash
& Gemini CLI
& 29.5 & \underline{23.0} & 6.5
& 28.0 & \underline{26.0} & 2.0
& 31.0 & \underline{20.0} & \textbf{11.0} \\
\bottomrule
\end{tabular}%
}
\endgroup
\end{table}

\begin{table}[t]
\centering
\caption{\textbf{Execution workload with and without task switching.}
Values are mean per-run totals; tokens are in millions.
Unint.\ covers the full uninterrupted session; Switch-All covers the full switched session; Switch-Focus covers resumed substantive focus stages, excluding the initial request and finalization.
Reasoning settings follow Table~\ref{tab:main_results}.}
\label{tab:resumption_resources}
\begingroup
\small
\setlength{\tabcolsep}{3.5pt}
\renewcommand{\arraystretch}{1.16}
\resizebox{\linewidth}{!}{%
\begin{tabular}{@{}llrrrrrrrrr@{}}
\toprule
\multirow{2}{*}{\textbf{Model}}
& \multirow{2}{*}{\textbf{Harness}}
& \multicolumn{3}{c}{\textbf{Tokens ($10^6$)}}
& \multicolumn{3}{c}{\textbf{Model calls}}
& \multicolumn{3}{c}{\textbf{Tool calls}} \\
\cmidrule(lr){3-5}
\cmidrule(lr){6-8}
\cmidrule(lr){9-11}
& & Unint. & \shortstack{Switch\\All} & \shortstack{Switch\\Focus}
  & Unint. & \shortstack{Switch\\All} & \shortstack{Switch\\Focus}
  & Unint. & \shortstack{Switch\\All} & \shortstack{Switch\\Focus} \\
\midrule

DeepSeek-V4.1-Flash
& OpenCode
& 8.594 & 101.953 & 25.375
& 112.2 & 402.2 & 120.3
& 115.2 & 405.5 & 124.7 \\

GLM-5.3
& OpenCode
& 4.904 & 14.970 & 12.511
& 94.5 & 163.7 & 94.5
& 102.8 & 181.4 & 103.1 \\

Kimi K3
& KimiCode
& 4.739 & 24.335 & 6.677
& 70.5 & 191.0 & 79.7
& 71.5 & 193.1 & 80.8 \\

GPT-5.6 Terra (medium)
& Codex
& 1.511 & 9.741 & 2.880
& 38.4 & 104.8 & 35.6
& 32.9 & 87.5 & 30.0 \\

GPT-5.6 Sol (medium)
& Codex
& 2.294 & 16.921 & 4.047
& 48.4 & 170.1 & 49.3
& 48.6 & 135.6 & 47.0 \\

Claude Opus 5 (medium)
& Claude Code
& 3.145 & 18.237 & 5.347
& 65.1 & 220.0 & 72.8
& 68.5 & 222.2 & 73.4 \\

Gemini 3.7 Flash
& Gemini CLI
& 8.880 & 46.058 & 15.840
& 57.8 & 245.9 & 28.5
& 122.8 & 320.8 & 117.5 \\

\bottomrule
\end{tabular}%
}
\endgroup
\end{table}

\subsection{Interruption Locality}
\label{sec:locality_results}

The Near and Far panels of Table~\ref{tab:main_results} compare interruptions within shared versus different immediate source directories (Section~\ref{sec:locality}).
Each group contains 100 distinct focus tasks, with uninterrupted and switched execution paired within each task.
\\\textbf{The far group consistently exhibits larger gaps.}
All seven configurations show larger resolution gaps in the far group.
The average gap is 8.1 \% for near interruptions and 14.9 \%  for far interruptions, with within-configuration differences ranging from 3.0 to 10.0 \% .
GLM-5.3 with OpenCode shows the largest difference: its gap increases from 1.0 \%  in the near group to 11.0 \%  in the far group.
Even the smallest difference, observed for Kimi K3, follows the same direction, with gaps of 12.0 and 15.0 \% .
The pattern is therefore consistent across the evaluated configurations despite variation in their overall resolution.
\\\textbf{Uninterrupted performance does not explain the locality pattern.}
All seven configurations achieve higher switched resolution in the near group.
For five configurations, this reverses the uninterrupted ordering: they perform better on far targets without interleaving but better on near targets under switching.
The larger far-group gaps therefore cannot be explained simply by lower uninterrupted resolution on those targets.
This pattern is consistent with nearby work providing more reusable context, although the results do not establish that mechanism.
Because the groups contain different focus tasks, the comparison indicates an association between source locality and resumption performance rather than an isolated causal effect of locality.

\subsection{Computation During Task Resumption}
\label{sec:resumption}

Table~\ref{tab:resumption_resources} reports cumulative processed tokens, model calls, and tool calls for uninterrupted sessions, full switched sessions, and resumed substantive focus stages.
Switch-Focus excludes the initial focus request, interruption work, and finalization.
\\\textbf{Switching increases session-level computation.}
Full switched sessions consume 4.5-11.1 times as many tokens as uninterrupted sessions, compared with 2.4-4.3 times as many model calls.
Token usage thus grows more than call counts for every configuration, indicating greater average context-processing volume per call.
These ratios describe the combined workload of focus and interruption tasks, rather than the isolated cost of resumption.
\\\textbf{Resumed focus work retain the cost of interruption history.}
Across seven configurations, resumed focus stages consume 1.41--2.95 times the tokens of a full uninterrupted session, without uniformly higher model-call counts.
Excluding interruption execution here does not exclude its history from subsequent focus-stage inputs.
Repeated processing of that history may increase token usage, although aggregate usage cannot separate this effect from differences in outputs and native context management.
This suggests value in distinguishing history needed for resumption from intervening interactions, while retaining the requirements and progress for continuing unfinished work.

\section{Context Management for Task Resumption}
\label{sec:context_management}

Section~\ref{sec:experiments} reports consistent resolution gaps under task switching despite persistent task-specific checkouts.
These checkouts preserve code changes, but resumption also requires recovering task-specific requirements and progress from the shared conversation.
We introduce a \emph{Resumption Context Manager} (RCM) that combines task-relevant history routing with explicit task-state memory.
As an OpenCode plug-in, RCM modifies the model's context while preserving the native execution workflow.
Figure~\ref{fig:rcm_overview} illustrates how RCM reconstructs task-specific contexts.

\subsection{Context Requirements for Resumption}
\label{sec:context_failure_analysis}

Trajectory inspection motivates two complementary recovery needs.
\\\textbf{Task-relevant history.}
Shared conversations mix requirements, observations, and decisions from several unfinished tasks.
On resumption, agents must identify relevant interactions and separate other tasks' assumptions.
The Moto trajectory illustrates this need at finalization, when it plans to collect changes across checkouts instead of submitting only the focus patch (Appendix~\ref{app:trajectory_moto}).
\\\textbf{Accumulated task state.}
Relevant history may distribute implemented requirements, earlier decisions, and pending work across many interactions.
The source tree does not explicitly capture this state.
In Click, a later extension leaves an earlier requirement unresolved (Appendix~\ref{app:trajectory_click}).
Resumption requires reconciling the cumulative specification with implementation progress.

\subsection{Resumption Context Manager}
\label{sec:context_module}

RCM combines a \emph{Task History Router} for selecting relevant interactions with \emph{Task State Memory} for recording progress and pending work.
For request $q_t$, available history $H_{<t}$, and inferred task set $Z_t$, the model receives
$
\mathcal{C}_t
=
\operatorname{Assemble}\!\left(
q_t,\,
\operatorname{Route}(H_{<t},Z_t),\,
\{M_z\}_{z\in Z_t}
\right),
$
where $M_z$ records task $z$'s state.
Multi-task requests use history and memory associated with all assigned tasks.
\\\textbf{Task History Router.}
An auxiliary model assigns each request to existing or new tasks using a task catalogue and earlier user messages.
Assignment considers semantic goals, explicit references, paths, and code symbols; the result is cached within the user turn.
The router retains interaction groups for active and related tasks, keeping user requests, assistant actions, and tool results together.
Shared instructions and unassigned interactions remain visible.
When routing confidence falls below 0.8, the router retains the full available history.
\\\textbf{Task State Memory.}
An auxiliary model incrementally updates each task's goal, constraints, progress, pending work, relevant files, and uncertainties from completed interactions.
Updates distinguish user requirements, tool observations, and assistant claims.
Constraints and progress retain source-message references checked for validity.
Long interactions are processed in bounded batches.
On resumption, memory accompanies the selected history.
Original user requests and recent interactions are preserved; older assistant and tool exchanges are replaced only when covered by an injected memory record.
Memory serves as historical context, subordinate to current user instructions.
\\\textbf{Integration.}
RCM modifies outgoing context through OpenCode's plug-in interface, preserving the stored conversation, task checkouts, and native model-tool loop.
Task assignments and memory persist within sessions.
Auxiliary calls default to the coding model.
For native compaction, RCM passes the full available input and instructs the compactor to preserve each task's requirements, progress, and pending work.
+Router and +RCM share routing and compaction behavior; +RCM additionally maintains and injects task-state memory.
Both use only agent-visible content.

\subsection{Evaluation and Results}
\label{sec:module_evaluation}
\label{sec:module_results}

We compare native OpenCode, +Router, and +RCM under uninterrupted and switched execution, holding task environments, model settings, execution budgets, and verification fixed within each model.
Each run starts with empty task assignments and memory.
Auxiliary routing and memory calls count toward the execution budget and total model usage.
\\\textbf{Task-aware context management improves switched completion.}
The +Router configuration raises switched resolution by 11.0 points for DeepSeek-V4.1-Flash and 3.5 points for GLM-5.3, while uninterrupted resolution changes by only $-1.0$ and $+0.5$ points.
These gains support task-aware routing and compaction as useful interventions under switching.
For GLM-5.3, the gain occurs entirely in the far group, where resolution rises from 40.0\% to 47.0\%.
\\\textbf{State memory adds gains beyond routing.}
Adding memory further improves switched resolution by 4.5 points for DeepSeek-V4.1-Flash and 2.0 points for GLM-5.3.
Full RCM yields larger gains over native OpenCode in the far group than in the near group for both models.
The overall uninterrupted-switched gaps narrow to $-1.0$ and $1.0$ points, while uninterrupted resolution remains within 0.5 points of native performance.
This pattern supports explicit task-state memory beyond routing, without establishing a benefit from switching itself.

\begin{table}[t]
\centering
\caption{\textbf{Context management improves switched completion on OpenCode.}
Overall/Near/Far contain 200/100/100 tasks.
Unint.\ and Switch report resolution (\%); Gap is their difference in \%.
Bold marks the highest Switch within each model and task group.}
\label{tab:module_results}
\begingroup
\small
\setlength{\tabcolsep}{6pt}
\renewcommand{\arraystretch}{1.16}
\resizebox{\linewidth}{!}{%
\begin{tabular}{@{}llrrrrrrrrr@{}}
\toprule
\multirow{2}{*}{\textbf{Model}}
& \multirow{2}{*}{\textbf{Harness}}
& \multicolumn{3}{c}{\textbf{Overall}}
& \multicolumn{3}{c}{\textbf{Near}}
& \multicolumn{3}{c}{\textbf{Far}} \\
\cmidrule(lr){3-5}
\cmidrule(lr){6-8}
\cmidrule(lr){9-11}
& & Unint. & Switch & Gap
  & Unint. & Switch & Gap
  & Unint. & Switch & Gap \\
\midrule

\rowcolor{black!5}
\multicolumn{11}{l}{\textbf{Native}} \\
DeepSeek-V4.1-Flash
& OpenCode
& \textbf{46.5} & 31.5 & 15.0
& 45.0 & 34.0 & 11.0
& \textbf{48.0} & 29.0 & 19.0 \\
GLM-5.3
& OpenCode
& 49.0 & 43.0 & 6.0
& 47.0 & 46.0 & \textbf{1.0}
& \textbf{51.0} & 40.0 & 11.0 \\

\addlinespace[4pt]
\rowcolor{black!5}
\multicolumn{11}{l}{\textbf{+Router}} \\
DeepSeek-V4.1-Flash
& OpenCode
& 45.5 & 42.5 & 3.0
& \textbf{47.0} & 42.0 & 5.0
& 44.0 & 43.0 & 1.0 \\
GLM-5.3
& OpenCode
& \textbf{49.5} & 46.5 & 3.0
& 49.0 & 46.0 & 3.0
& 50.0 & 47.0 & 3.0 \\

\addlinespace[4pt]
\rowcolor{black!5}
\multicolumn{11}{l}{\textbf{+RCM}} \\
DeepSeek-V4.1-Flash
& OpenCode
& 46.0 & \textbf{47.0} & $\mathbf{-1.0}$
& 46.0 & \textbf{46.0} & \textbf{0.0}
& 46.0 & \textbf{48.0} & $\mathbf{-2.0}$ \\
GLM-5.3
& OpenCode
& \textbf{49.5} & \textbf{48.5} & \textbf{1.0}
& \textbf{51.0} & \textbf{49.0} & 2.0
& 48.0 & \textbf{48.0} & \textbf{0.0} \\
\bottomrule
\end{tabular}%
}
\endgroup
\end{table}

\section{Conclusion}
\label{sec:conclusion}

We introduce \ourmethod to evaluate unfinished-task resumption through matched uninterrupted and switched sessions.
Across 200 focus tasks from 19 repositories, the benchmark preserves requirements, verifiers, and task-specific workspaces.
Seven model--agent configurations show an average resolution gap of 11.5\% under task switching within one repository, indicating that code persistence alone does not ensure reliable resumption.
Our Resumption Context Manager combines task-relevant history routing with explicit task-state memory, improving switched resolution by 5.5--15.5\% on OpenCode while maintaining comparable uninterrupted performance.

\section*{AI use statement}
In this work, we used generative AI models as the coding agents evaluated in our benchmark and as auxiliary components of our proposed Resumption Context Manager. Their roles, configurations, and evaluation procedures are described in the main text and appendix.
We also used generative AI tools for assistance with
interpreting experimental results and debugging code used in our research implementation.
We have not used generative AI tools to generate synthetic benchmark datasets, clean or reformat benchmark data, or perform qualitative classification of experimental trajectories, and the formulation and proof of mathematical claims are not applicable to this work.
Additionally, we used generative AI tools to identify and summarize related literature, revise manuscript text, and develop and refine plotting scripts and scientific figures.
These uses are separate from the language models evaluated as coding agents and the auxiliary model calls used by our proposed method, which are described in the paper.
We have reviewed all AI-assisted work. Specifically, we reviewed the manuscript text, references, code changes, figures, and interpretations of the experimental results, and retained responsibility for all research decisions and conclusions.
We take responsibility for the final content of this work, including text, claims or artifacts produced with the aid of generative AI.

\section*{Ethics statement}

This work evaluates coding agents using repository repair tasks
and development histories from public software projects. Its
purpose is to characterize the reliability of agents when resuming
unfinished work, rather than to assess the performance of individual
developers or repository contributors. 

\section*{Reproducibility statement}

The paper documents the benchmark and evaluation procedures needed
to reproduce our study. Section~\ref{sec:benchmark} defines task
construction, matched uninterrupted and switched schedules,
source-directory locality, and executable verification;
Appendix~\ref{app:benchmark_construction} provides additional
construction and schedule-validation details.
Appendix~\ref{sec:experimental_setup} describes the evaluated
model--agent configurations, execution environments, request
interface, scoring protocol, and resource accounting.
Section~\ref{sec:context_module} and
Appendix~\ref{app:rcm_details} describe the Resumption Context
Manager, with its evaluation presented in
Section~\ref{sec:module_evaluation}.
We plan to publicly release the benchmark data and code to support
independent reproduction and further analysis. 

\bibliography{iclr2027_conference}
\bibliographystyle{iclr2027_conference}

\appendix
\clearpage
\section{Comparison with Related Benchmarks}
\label{sec:app_comparison}

Table~\ref{tab:benchmark_comparison} compares \ourmethod{}
with related benchmarks in terms of dataset profiles and
interaction capabilities.
The key distinction is the interleaving and repeated resumption
of unfinished coding tasks, each with a persistent workspace.

\providecolor{paBlue}{RGB}{232,240,254}
\providecolor{paGreen}{RGB}{236,253,245}
\providecolor{paYellow}{RGB}{255,251,235}
\providecolor{paPink}{RGB}{255,240,246}

\begin{table}[t]
\centering
\setlength{\tabcolsep}{3.2pt}
\renewcommand{\arraystretch}{1.14}
\caption{\textbf{Dataset profiles and capabilities of nearby agent benchmarks}.
The capability headings denote multi-stage interaction (MSI), separate coding tasks (SCT), interleaved unfinished tasks (IFT), repeated task resumption (RTR), and persistent per-task workspaces (PPTW).
Check marks indicate support, circles partial or adjacent coverage, and crosses no support.
}
\label{tab:benchmark_comparison}
\resizebox{\linewidth}{!}{%
\begin{tabular}{lccc@{\hspace{7pt}}ccccc}
\toprule
& \multicolumn{3}{c}{\textbf{Dataset profile}}
& \multicolumn{5}{c}{\textbf{Benchmark capabilities}} \\
\cmidrule(lr){2-4}\cmidrule(lr){5-9}
\shortstack[l]{Benchmark name}
& \shortstack{Size}
& \shortstack{Repos / sites}
& \shortstack{Primary scope}
& \shortstack{MSI}
& \shortstack{SCT}
& \shortstack{IFT}
& \shortstack{RTR}
& \shortstack{PPTW} \\
\midrule
SWE-bench~\citep{jimenez2024swebench}
& 2,294 & 12 & \shortstack{Issue repair}
& \xmark & \xmark & \xmark & \xmark & \xmark \\
SWE-bench-Live~\citep{zhang2025swebenchlive}
& 1,319 & 93 & \shortstack{Fresh issue repair}
& \xmark & \xmark & \xmark & \xmark & \xmark \\
SWE-ContextBench~\citep{zhu2026swecontextbench}
& $300{+}99$ & 12 & \shortstack{Context reuse}
& \pmark & \cmark & \xmark & \xmark & \xmark \\
InterruptBench~\citep{zou2026interruptbench}
& 165 & 5 & \shortstack{Web goal changes}
& \cmark & \xmark & \pmark & \pmark & \xmark \\
SWE-Touch~\citep{tan2026swetouch}
& 500 & 12 & \shortstack{User code edits}
& \cmark & \xmark & \pmark & \pmark & \xmark \\
\midrule
\rowcolor{paBlue}
\textbf{Switch-SWE (ours)}
& \textbf{200} & \textbf{19} & \shortstack{\textbf{Task} \textbf{switching}}
& \cmark & \cmark & \cmark & \cmark & \cmark \\
\bottomrule
\end{tabular}%
}
\end{table}

\section{Experimental Setup}
\label{sec:experimental_setup}

We evaluate \ourmethod{} through the native coding-agent interfaces used in the main experiments, while Harbor provides a common execution and verification layer. Each run preserves one continuous agent session and five persistent task-specific checkouts. The uninterrupted and switched conditions are executed independently from clean initial states; only the final focus-task implementation is scored.

\subsection{Agent Systems}
\label{sec:agent_systems}

Table~\ref{tab:agent_systems} lists the evaluated harnesses, model backbones, and reasoning settings. We use the highest reasoning effort exposed by a harness unless a medium setting is stated explicitly. Agent and harness versions are fixed for the full benchmark sweep.

\subsection{Runtime Environment}
\label{sec:runtime_environment}

\paragraph{Harbor execution.}
All evaluations run through Harbor. Each session is provisioned with 4 vCPU, 16\,GiB RAM, and 30\,GiB disk, together with the task-specific dependencies required by the source repository. The coding harness runs in the evaluation environment and communicates with its configured model endpoint through its native interface. External issue trackers and repository-hosting interfaces are not used to recover task context during a run.

\paragraph{Persistent task workspaces.}
Every run materializes five role-neutral checkouts, one for the focus task and four for the tasks that may appear as interruptions. Each checkout starts from its task's pinned base revision and persists for the full session. Switching changes the active checkout but does not reset files from previously visited tasks. Uninterrupted controls materialize the same five-checkout layout, although only focus requests are delivered. The paired conditions share no modified files, conversation state, or agent memory.

\paragraph{Continuous agent session.}
All requests in a run are delivered to one continuous native agent conversation. Native context management and compaction remain enabled according to the evaluated harness configuration. The controller does not create a new conversation when the active task changes and does not provide a task-specific recovery summary on return.

\paragraph{Per-request execution budget.}
Every scheduled request, including finalization, receives a 900-second wall-clock budget per attempts and for 3 times retry. A request ends when the agent yields or the deadline is reached; any code already written to the active checkout remains available to later requests. The controller then advances to the next scheduled request. 

\paragraph{Infrastructure retries.}
Retries are reserved for confirmed infrastructure faults that occur outside the agent's substantive work, such as environment-provisioning or verifier-service failures. At most two such retries are allowed. Agent timeouts, invalid patches, failed tests, and model-side errors are not converted into additional attempts. When verification alone must be retried, the same frozen focus artifact is reused.

\subsection{Request Interface}
\label{app:release_schema}

The controller sends one request at a time. Each message identifies the repository, the active checkout path, and the current natural-language requirement. The same focus-task messages, including finalization, are used in the uninterrupted and switched conditions. The active path is explicit on every request so the agent can operate on the intended checkout without receiving private focus/interruption labels.

Requests expose only information available at that point in the schedule. The controller withholds future requests, reference implementations, held-out scoring tests, private locality labels, and verifier feedback. On a return to an earlier task, the agent receives the new request and active checkout path but no controller-generated summary of earlier work. This makes recovery from the accumulated conversation and persistent source tree part of the evaluated capability.

The finalization request adds no new behavioral requirement. It asks the agent to finish the active focus task and prepare the implementation for evaluation. Only the focus checkout is collected for final scoring.

\begin{table}[t]
\centering
\caption{\textbf{Evaluated harnesses and models.}
One row denotes one native coding harness.}
\label{tab:agent_systems}
\begingroup
\small
\setlength{\tabcolsep}{5pt}
\renewcommand{\arraystretch}{1.15}
\begin{tabular*}{\linewidth}{@{\extracolsep{\fill}}lll@{}}
\toprule
\textbf{Harness} & \textbf{Models and reasoning effort} & \textbf{Version} \\
\midrule
OpenCode
& DeepSeek-V4.1-Flash, GLM-5.3
& 1.18.2 \\
KimiCode
& Kimi K3
& 0.42.0 \\
Codex
& GPT-5.6 Terra, GPT-5.6 Sol
& 0.154.0 \\
Claude Code
& Claude Opus 5
& 2.1.269 \\
Gemini CLI
& Gemini 3.7 Flash
& 0.60.0 \\
\bottomrule
\end{tabular*}
\endgroup
\end{table}

\subsection{Evaluation Protocol}
\label{sec:evaluation_protocol}

\paragraph{Matched evaluation.}
Each of the 200 focus tasks is evaluated once in the uninterrupted condition and once in the switched condition. The 200 tasks comprise 100 Near and 100 Far targets. Within a matched pair, the focus-task base revision, focus request sequence, checkout path, per-request budget, finalization message, and verifier are identical. The two runs are independent and begin from clean state.

\paragraph{Final focus artifact.}
After finalization, the evaluator parse the agent's submitted focus-task patch and packaging normalization. The artifact includes committed and uncommitted source changes, deletions, and newly created project files. Local environments, harness logs, caches, and other execution artifacts are excluded. Changes in interruption checkouts are retained for analysis but never enter the focus-task score.

\paragraph{Clean verification.}
The verifier reconstructs the pinned focus base revision in a clean environment, applies the frozen focus artifact, adds the held-out scoring-test patch, and runs the fixed task-specific command. Agent-authored tests may be used during development but do not replace the held-out verifier. This follows the executable final-patch evaluation used by SWE-bench-style repository-repair benchmarks~\citep{jimenez2024swebench}.

Let $\mathcal{F}_f$ and $\mathcal{P}_f$ denote the designated fail-to-pass and pass-to-pass tests for focus task $f$. The task is resolved iff the artifact applies successfully and every required test is collected, executed, and passes:
\begin{equation}
\operatorname{Resolved}(f)
=
\mathbb{I}\!\left[
\operatorname{Apply}(f)=1
\;\land\;
\bigwedge_{u\in\mathcal{F}_f\cup\mathcal{P}_f}
\operatorname{Pass}(u)=1
\right].
\label{eq:appendix_resolution_criterion}
\end{equation}
A missing finalization, an uncollectable or inapplicable artifact, a missing/skipped/failing required test, or verifier timeout is unresolved. Confirmed infrastructure faults may trigger re-execution of the same frozen artifact under the retry policy above; they do not trigger a new agent rollout.

\subsection{Metrics and Resource Accounting}
\label{sec:metrics}

\paragraph{Resolution and resumption gap.}
For a set $G$ of focus tasks, resolution is the fraction whose final focus artifact satisfies Equation~\ref{eq:appendix_resolution_criterion}. The denominator is fixed by the benchmark split: $|G|=200$ for Overall and $|G|=100$ for each locality group. Runs that terminate early or never produce a resolvable final artifact remain in the denominator as unresolved. For a model--harness configuration $m$, the reported resumption gap is
\begin{equation}
\operatorname{Gap}_m(G)
=
\operatorname{Res}^{\mathrm{unint}}_m(G)
-
\operatorname{Res}^{\mathrm{switch}}_m(G),
\label{eq:gap_metric}
\end{equation}
reported in percentage points.

\paragraph{Trajectory-level computation.}
Following the trajectory-level accounting used in SWE-Marathon~\citep{desai2026swemarathon}, we report cumulative token processing, model calls, and tool calls. Tokens are summed across model invocations as cumulative input plus output usage. Cached input is included when reported by the provider, and overlapping aggregate/provider fields are counted once. If reasoning tokens are already included in output usage, they are not added a second time. Reprocessing the same context in a later model call contributes again because it consumes another inference step.

A \emph{model call} is one native model-inference step issued by the harness. A \emph{tool call} is one executed tool invocation recorded in the agent trajectory. These counts are distinct from scheduled user requests and from Harbor orchestration events.

For switched sessions, \textbf{All} covers the entire run, including interruption work and finalization. \textbf{Focus} covers only substantive focus requests that occur after intervening work; it excludes the initial focus request, every interruption request, and finalization. Runs that end early contribute the computation actually incurred before termination. Table~\ref{tab:resumption_resources} reports the mean per switched session for each configuration.

\section{Benchmark Construction and Schedule Validation}
\label{app:benchmark_construction}

\ourmethod{} combines executable repository repairs with ordered, cumulative requirements reconstructed from development histories. Construction is designed so that the same final focus behavior can be evaluated under uninterrupted and interleaved sessions without changing the task's verifier.

\subsection{Executable Source Instances}
\label{app:source_instances}

We source repair instances from the SWE-bench family and additional merged pull requests in Python repositories~\citep{jimenez2024swebench,zhang2025swebenchlive}. Each retained task has a pinned base revision, traceable issue or pull-request material, a reference implementation, and a reproducible execution environment. The implementation patch and test patch are separated so the same scoring tests can be run against the base state and the reference repair.

The benchmark manifest records the source repository and revision, delivered requests, environment recipe, designated test identifiers, and the verifier command. Reference implementations and held-out scoring assets are kept outside agent-visible workspaces.

\subsection{Test Classification and Executable Qualification}
\label{app:test_classification}

For each task, the designated tests are run on the base revision with the scoring-test patch and on the same state after applying the reference implementation. A test is \emph{fail-to-pass} (F2P) when it fails on the base state and passes with the reference repair. A test is \emph{pass-to-pass} (P2P) when it passes in both states. Retained tasks contain at least one F2P test and designated P2P coverage for regression protection.

A required test contributes to a successful resolution only when it is collected, executed, and passes. Missing, skipped, failing, or timed-out required tests do not satisfy the criterion. The same designated F2P/P2P sets are used for the uninterrupted and switched member of every matched pair.

\subsection{Requirement Reconstruction and Review}
\label{app:requirement_reconstruction}

\paragraph{Initial requirement.}
The first substantive request is reconstructed from the linked issue title and body, or from the pull-request description when no linked issue is available. We preserve the requested behavior, public interfaces, and user-visible constraints while removing reference-solution code and repository-maintainer details that are not needed to state the task.

\paragraph{Follow-up requirements.}
Later requests follow substantive pre-merge review rounds chronologically. Comments that express one coherent change in expected behavior are consolidated into one request. Automated messages, duplicate comments, approvals without new requirements, and progress-only discussion are excluded. A follow-up is retained only when it remains interpretable from the earlier requests and the agent-visible repository, without access to an unavailable intermediate maintainer patch.

Where additional behavioral coverage is needed, benchmark-authored extensions are stated as ordinary user-visible requirements and paired with regression tests. Their provenance is recorded separately from upstream issue or review text. Regardless of source, requirements are cumulative: a later request extends the task rather than replacing earlier accepted behavior.

\paragraph{Requirement--verifier alignment.}
We audit the complete request sequence against the scoring tests. Every F2P behavior must be justified by a delivered requirement or an explicitly stated public contract, and later requests must preserve requirements established by earlier stages. Cases whose intended behavior cannot be stated cleanly or verified reproducibly are excluded. The resulting metadata links source material to requests and requests to executable checks.

\subsection{Interruption Selection}
\label{app:interruption_selection}

Every switched session pairs its focus task with four distinct unfinished tasks from the same repository. Each task keeps a separate checkout, so interruption work cannot overwrite the focus task's files. The shared conversation is therefore the principal cross-task state channel.

Locality follows the source-directory definition in Section~\ref{sec:locality}. Interruption tasks are source-file-disjoint from the focus task. In the Near group, each focus--interruption pair modifies distinct source files but shares at least one immediate source directory; in the Far group, the corresponding source-directory sets are disjoint. The benchmark contains 100 focus tasks in each group. A focus task appears in only one group, while an eligible source task may serve as an interruption in more than one session.

\subsection{Schedule Generation and Validation}
\label{app:schedule_generation}

The scheduler constructs the switched session before agent execution. For each of the four interruption tasks, it selects an ordered prefix of substantive requests and divides that prefix into one or more non-contiguous chunks. These chunks are inserted between focus requests while preserving the original request order within every task. Consequently, each interruption task is suspended and resumed at least once, and the focus task returns multiple times before finalization.

The uninterrupted control is the exact projection of the switched schedule onto focus requests. Automated validation checks the following properties before execution:
\begin{itemize}
    \item the focus-message payloads and their order are identical across the matched pair;
    \item every task preserves its own request order;
    \item exactly four distinct same-repository interruption tasks are used;
    \item each interruption satisfies the assigned locality and source-file-disjointness constraints;
    \item the active checkout path is valid for every request; and
    \item focus finalization is the final scheduled request.
\end{itemize}
The released session specification records the full request schedule, task-to-workspace mapping, locality assignment, and scheduler seed. Task selection and schedule assembly are fixed before the reported model evaluations.
Table~\ref{tab:session_stats} summarizes the benchmark scale
and the resulting switched-session schedules.

\begin{table}[!htbp]
\centering
\caption{\textbf{Benchmark profile.}
Each focus task yields one matched pair of uninterrupted and
switched sessions.
Focus-task and repository counts are dataset totals;
the interruption count is the number of distinct interruption
tasks per switched session.
Request, workspace-switch, and focus-return counts are medians
over switched sessions; requests include finalization.}
\label{tab:session_stats}
\begingroup
\small
\setlength{\tabcolsep}{4pt}
\renewcommand{\arraystretch}{1.15}
\begin{tabular*}{\linewidth}{@{\extracolsep{\fill}}rrrrrr@{}}
\toprule
\shortstack[r]{\textbf{Focus}\\\textbf{tasks}}
& \textbf{Repos.}
& \shortstack[r]{\textbf{Interruption}\\\textbf{tasks}}
& \textbf{Requests}
& \shortstack[r]{\textbf{Workspace}\\\textbf{switches}}
& \shortstack[r]{\textbf{Focus}\\\textbf{returns}} \\
\midrule
200 & 19 & 4 & 22 & 15 & 3 \\
\bottomrule
\end{tabular*}
\endgroup
\end{table}

\section{Illustrative Task Catalog}
\label{app:task_catalog}

Following the task-level presentation used by long-horizon coding benchmarks, we describe representative \ourmethod{} tasks by their cumulative objective and executable verifier. Table~\ref{tab:task_catalog} summarizes five examples spanning library behavior, process and service interactions, and graph algorithms. The examples are selected to make the resumption requirement concrete rather than to define separate benchmark categories.

\begin{table}[htbp]
\centering
\caption{\textbf{Illustrative \ourmethod{} tasks.}
F2P and P2P are the designated fail-to-pass and pass-to-pass tests used by the final focus verifier.}
\label{tab:task_catalog}
\begingroup
\small
\setlength{\tabcolsep}{4pt}
\renewcommand{\arraystretch}{1.16}
\begin{tabular}{@{}p{0.16\linewidth}p{0.52\linewidth}r@{}}
\toprule
\textbf{Repository} & \textbf{Cumulative objective} & \textbf{F2P / P2P} \\
\midrule
Packaging
& Restore serialized tags across supported state formats while preserving an existing object's identity and hash when restoration fails.
& 13 / 217 \\
Click
& Reconcile flag activation with literal and callable defaults while preserving Boolean values and callable flag values.
& 7 / 732 \\
AnyIO
& Compose child-session creation with validated, single-pass file-descriptor inheritance.
& 10 / 18 \\
Moto
& Preserve S3--SNS--SQS raw delivery while selecting the subscription-specific payload before message-body filtering.
& 5 / 8 \\
NetworkX
& Construct planar dual multigraphs and contract selected dual edges without losing original face provenance.
& 25 / 41 \\
\bottomrule
\end{tabular}
\endgroup
\end{table}

\subsection{Packaging: Failure-Atomic Tag Restoration}
\label{app:catalog_packaging}

\paragraph{Objective.}
Extend \texttt{packaging.tags.Tag} so serialized tags can be restored safely. The interpreter, ABI, and platform determine tag identity, while the cached hash must be recomputed after restoration rather than treated as portable serialized state.

\paragraph{Cumulative requests.}
The first request adds compact serialization while retaining compatibility with existing tuple and slot-state forms. A follow-up accepts three restoration representations: a three-string tuple, a dictionary containing the three fields, and the legacy \texttt{(None, slot\_dict)} form. Accepted strings undergo the same normalization as ordinary construction. The final requirement adds failure atomicity: if normalization or hashing raises, an already initialized tag must retain its previous string representation, equality behavior, and hash.

\paragraph{Verifier.}
The final suite contains 13 F2P and 217 P2P tests. It covers real pickle \texttt{BUILD} restoration, multiple pickle protocols, malformed inputs, string-subclass failures, and repeated failed restorations. Passing requires both the newly requested restoration behavior and the existing tag behavior to remain correct.

\paragraph{State to recover.}
A resumed agent must preserve all earlier accepted state formats and normalization rules while adding the later atomicity guarantee.

\subsection{Click: Literal Values and Callable Defaults}
\label{app:catalog_click}

\paragraph{Objective.}
Reconcile two meanings of a flag's \texttt{default}: passing a Python value directly to the callback and activating the flag's \texttt{flag\_value}. For a non-Boolean flag, \texttt{True} can act as an activation sentinel; a Boolean flag must retain the literal Boolean value.

\paragraph{Cumulative requests.}
A follow-up extends the rule to default factories. A factory evaluated by \texttt{get\_default(ctx, call=True)} is invoked once, and a \texttt{True} result follows the non-Boolean activation rule. With \texttt{call=False}, the factory itself is returned without invocation. A later request protects the boundary cases: Boolean defaults remain literal, a callable \texttt{flag\_value} remains a value rather than a factory, and a factory returning \texttt{None} continues to return \texttt{None}.

\paragraph{Verifier.}
Seven F2P and 732 P2P tests cover literal defaults, interactive Boolean prompts, callable defaults, and callable flag values. Literal and factory-produced \texttt{True} are tested separately.

\paragraph{State to recover.}
The resumed agent must retain the distinction between a callable default factory and a callable flag value, while applying the same cumulative reconciliation rule to both literal and factory-produced defaults.

\subsection{AnyIO: Session Creation and Descriptor Handoff}
\label{app:catalog_anyio}

\paragraph{Objective.}
Extend \texttt{run\_process()} and \texttt{open\_process()} so child-process creation supports both a new session and explicit file-descriptor inheritance without changing ordinary process I/O and cleanup behavior.

\paragraph{Cumulative requests.}
The first request adds \texttt{start\_new\_session=False} and forwards it through the supported execution paths. The next introduces \texttt{pass\_fds=()}: its iterable is consumed once, duplicate descriptors are tolerated, and invalid descriptors are rejected before child creation. The final request requires the two features to compose, so a child can inherit a pipe while entering a new session without closing the parent's descriptor.

\paragraph{Verifier.}
Ten F2P and 18 P2P tests check session identifiers, descriptor validation before launch, pipe inheritance, empty descriptor sets, and pre-existing process behavior across the supported public APIs.

\paragraph{State to recover.}
The descriptor-handoff extension must preserve the earlier session-creation behavior and work when both options are active in the same child launch.

\subsection{Moto: Selecting the Payload Before Filtering}
\label{app:catalog_moto}

\paragraph{Objective.}
Repair S3 notifications delivered through SNS to an SQS subscription with \texttt{RawMessageDelivery=true}. S3-originated messages may contain no message attributes, and the raw-delivery path must handle that case without dropping the notification.

\paragraph{Cumulative requests.}
A later requirement extends delivery to structured SNS messages. With \texttt{MessageStructure=json}, the subscription's \texttt{sqs} payload, or its \texttt{default} fallback, is selected before applying a \texttt{MessageBody} filter. A matching raw payload is preserved exactly, including whitespace and Unicode; non-raw delivery places the same selected payload inside its envelope. Attribute-scope filters, fanout isolation, and the original missing-attribute behavior must remain correct.

\paragraph{Verifier.}
Five F2P and eight P2P tests cover the S3--SNS--SQS regression, protocol-payload selection, default fallback, exact body preservation, and rejection when only the outer routing wrapper matches a filter.

\paragraph{State to recover.}
The resumed agent must retain the original null-handling behavior while adding the later ordering constraint that selects the protocol-specific payload before body filtering.

\subsection{NetworkX: Dual Graphs with Face Provenance}
\label{app:catalog_networkx}

\paragraph{Objective.}
Add face enumeration and dual-graph construction to \texttt{PlanarEmbedding}. Face boundaries are component-local walks: repeated vertices around bridges matter, isolated vertices create no walks, and disconnected components do not share one global outer face.

\paragraph{Cumulative requests.}
The first request establishes \texttt{faces()} and its half-edge partition. The next adds \texttt{dual\_graph()}, returning a multigraph with one node per face and one edge per undirected primal edge, while retaining parallel edges, bridge-induced loops, and maps from half-edges and original faces to the dual. The final request adds \texttt{contract\_edges}: selected dual edges are contracted while original face provenance remains available. Repeated or reversed selections are idempotent, arbitrary hashable node labels are supported, and the input embedding remains unchanged.

\paragraph{Verifier.}
Twenty-five F2P and 41 P2P tests check facial walks, triangle and tree duals, disconnected components, transitive contractions, retained loops, invalid inputs, heterogeneous node labels, and structural independence of the returned graph.

\paragraph{State to recover.}
Contraction changes the current dual regions but not the identity of the original faces. A resumed agent must preserve that distinction together with the original incidence maps established by earlier requests.

\section{Task-Resumption Trajectories}
\label{app:trajectory_case_studies}
\label{app:failure_modes}

The main benchmark score is determined only by the executable final focus verifier. Trajectory inspection serves a different role: it shows what information an agent must recover when it returns to an unfinished task. We examine the delivered requests, native conversation, tool trace, focus artifact, and verifier outcome together. The cases below illustrate the two recovery needs that motivate Section~\ref{sec:context_management}: keeping the active task isolated from other work and retaining the accumulated state of that task.

\subsection{Trace Signatures}
\label{app:failure_taxonomy}

We use two evidence-based signatures when reading switched trajectories. \emph{Requirement-state loss} is present when a later focus edit or plan conflicts with an earlier focus requirement or overlooks previously established implementation state. \emph{Cross-task interference} is present when requirements, files, or submission scope from another active task enter the focus task's implementation or finalization. We assign these signatures only when the relevant request, action, or submitted code is visible in the saved trace; the final verifier outcome alone is not sufficient.

\begin{table}[t]
\centering
\caption{\textbf{Trajectory signatures used in the case studies.}
The signatures describe observable resumption behavior and are separate from the binary resolution metric.}
\label{tab:failure_modes}
\begingroup
\small
\setlength{\tabcolsep}{5pt}
\renewcommand{\arraystretch}{1.14}
\begin{tabular}{@{}p{0.26\linewidth}p{0.64\linewidth}@{}}
\toprule
\textbf{Signature} & \textbf{Trace evidence} \\
\midrule
Requirement-state loss
& A returned focus edit or plan violates an earlier focus requirement, forgets a prior decision, or fails to preserve an invariant established before interruption. \\
Cross-task interference
& Work associated with another checkout enters the focus task's reasoning, edits, or final-submission scope. \\
\bottomrule
\end{tabular}
\endgroup
\end{table}

\subsection{Reconciling Earlier and Later Requirements}
\label{app:trajectory_click}

\noindent\textbf{Click; DeepSeek-V4.1-Flash with OpenCode.}
The task combines literal-default reconciliation with a callable-default extension (Appendix~\ref{app:catalog_click}). Both conditions use the same base revision and the same seven F2P and 732 P2P tests. The uninterrupted artifact passes every required test; the switched artifact passes three of seven F2P tests and all 732 P2P tests.

\paragraph{Uninterrupted execution.}
The agent first restricts \texttt{True}-to-\texttt{flag\_value} conversion to non-Boolean flags. It then implements default factories by evaluating the factory once and applying the same conversion rule, so literal and factory-produced \texttt{True} share one guard.

\paragraph{Switched execution.}
Before interruption, the agent treats the initial reconciliation as already implemented. After returning for the callable-default extension, it moves factory evaluation before activation resolution and adds the Boolean guard inside the callable branch, while the literal branch remains unchanged. The resulting implementation contains the following asymmetry:
\begin{quote}
\begin{minipage}{\linewidth}
\scriptsize
\begin{verbatim}
if call and callable(value):
    value = value()
    if value is True and self.is_flag and not self.is_bool_flag:
        value = self.flag_value
elif value is True and self.is_flag:
    value = self.flag_value
\end{verbatim}
\end{minipage}
\end{quote}
The final branch still converts literal \texttt{True} for Boolean flags with \texttt{flag\_value=False}.

\paragraph{What the trace shows.}
The switched agent successfully adds the later callable-default behavior but does not reconcile it with the earlier literal-default rule. Its local checks pass while the held-out verifier rejects the unguarded literal case and related Boolean-prompt cases. This is a direct example of requirement-state loss: the new requirement is implemented on top of an incomplete representation of the earlier task state.

\subsection{Maintaining Focus Identity Through Finalization}
\label{app:trajectory_moto}

\noindent\textbf{Moto; GPT-5.6 Sol with Codex.}
The focus task is the SNS-to-SQS delivery repair described in Appendix~\ref{app:catalog_moto}. After work on four other checkouts, the agent returns to the correct focus workspace and implements protocol-payload selection before body filtering.

\paragraph{Correct return, confused submission scope.}
During the returned implementation stage, the agent edits the correct checkout and handles missing message attributes, raw and non-raw delivery, and exact payload preservation. At finalization, however, it broadens the scope of its submission procedure and inspects diffs from all five workspaces:
\begin{quote}
\begin{minipage}{\linewidth}
\scriptsize
\begin{verbatim}
for d in /work/w00 /work/w01 /work/w02 /work/w03 /work/w04; do
    git -C "$d" diff
done | wc -l
\end{verbatim}
\end{minipage}
\end{quote}
The uninterrupted run produces a focus artifact that passes all five F2P and eight P2P tests, whereas the switched run does not produce an applicable final focus patch.

\paragraph{What the trace shows.}
Implementation recovery and submission recovery are separable. The agent returns to the right code and performs the intended repair, yet other active tasks re-enter its notion of what should be finalized. This is the cross-task interference pattern targeted by task-specific history routing.

\subsection{Preserving Invariants Across Returns}
\label{app:trajectory_networkx}

\noindent\textbf{NetworkX; DeepSeek-V4.1-Flash with OpenCode.}
Both conditions pass all 25 F2P and 41 P2P tests for the planar-dual task in Appendix~\ref{app:catalog_networkx}.

The switched agent first implements component-local face walks. After intervening work, it builds \texttt{dual\_graph()} on that representation and records original-face and half-edge incidence maps. On a later return, it adds contractions while preserving those maps as provenance. Its local checks include contracting one triangle dual edge while retaining the other two as loops.

\paragraph{What the trace shows.}
Successful resumption is not simply remembering the latest request. The agent extends an earlier representation while preserving the invariants established by previous stages. This case motivates retaining compact task state---requirements, decisions, completed work, and pending work---alongside the task-relevant conversation history.

\section{Resumption Context Manager: Implementation Details}
\label{app:rcm_details}

The Resumption Context Manager (RCM) in
Section~\ref{sec:context_module} operates at the context boundary
of OpenCode. It does not modify repository state, tool semantics,
or the native model--tool loop. Its two components---the Task
History Router and Task State Memory---use only content already
visible to the agent.

\begin{figure}[!htbp]
    \centering
    \includegraphics[width=\linewidth]{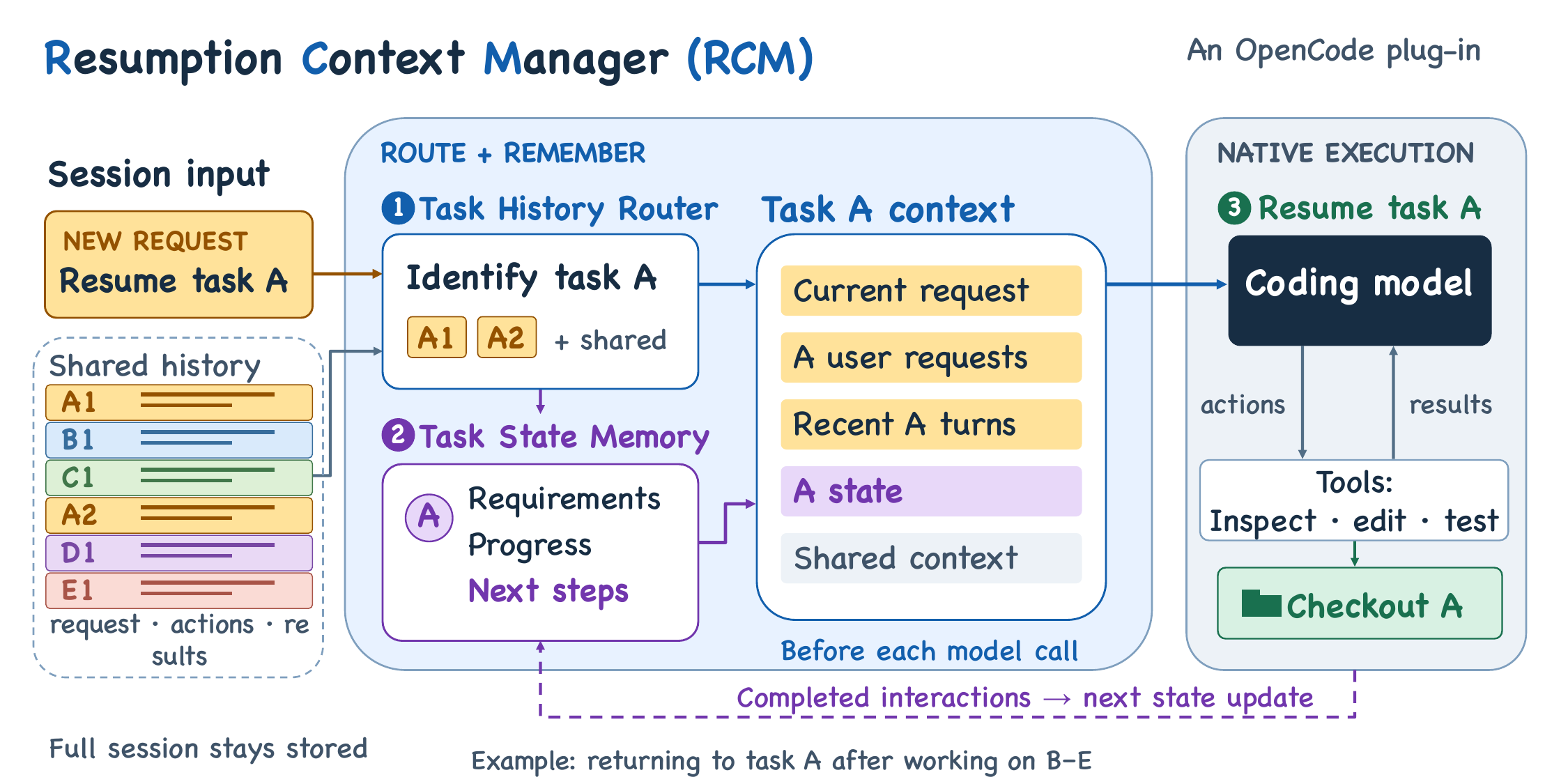}
    \caption{
    \textbf{Resumption Context Manager (RCM).}
    When task $A$ resumes, the Task History Router selects
    relevant interactions from the shared history, while
    Task State Memory supplies accumulated requirements,
    progress, and pending work.
    These signals are combined with the current request and
    shared context for the coding model.
    Completed interactions inform subsequent state updates;
    the native model--tool loop and task checkout are preserved.
    }
    \label{fig:rcm_overview}
\end{figure}

Figure~\ref{fig:rcm_overview} illustrates how the plugin reconstructs
context when an unfinished task resumes.
The plugin preserves the stored session history and changes only
the messages supplied to the current model call, combining
task-relevant interactions with shared context and, when available,
a compact task-state record.

\subsection{Task History Router}

For each incoming user request, the router assigns the request to an existing task or creates a new task entry. The assignment considers the request's semantic goal together with explicit task references, active paths, repository names, and code symbols present in the conversation. The assignment is cached for the turn.

History is stored as interaction groups so that a user request, the assistant actions taken in response, and their tool results remain together. When a task resumes, RCM selects the groups associated with that task and any explicitly related task. Shared instructions and unassigned interactions remain visible. If assignment confidence is below 0.8, RCM retains the full available history rather than aggressively filtering context.

\subsection{Task State Memory}

Task State Memory maintains a concise record for each task. The record contains the current goal, cumulative constraints, completed work, pending work, relevant files or symbols, and unresolved uncertainties. Updates are grounded in the conversation: user requests define requirements, tool outputs provide observed repository state, and assistant statements are retained as claims unless supported by subsequent tool evidence.

Memory is updated incrementally from completed interactions. On resumption, the task-state record is injected together with the routed history. Current user instructions always take precedence over stored state. Original user requirements and recent interactions remain verbatim; older assistant/tool exchanges are replaced only when the injected memory covers the information they contribute.

\subsection{Compaction and Evaluation}

Native OpenCode compaction remains enabled.
During compaction, the plugin bypasses task-specific history
filtering and adds an instruction to preserve the goals,
constraints, observed progress, and unresolved work of all topics,
including suspended ones.
The +Router condition uses the same routing and compaction
behavior but omits Task State Memory; +RCM enables both components.

The module evaluation uses the same task environments, model settings, 900-second per-request budget, final focus verifier, and resolution metric as the native OpenCode runs. Each run starts with empty task assignments and memory. Auxiliary routing and memory calls count toward the same execution budget and model-usage accounting as coding calls, so the reported switched-resolution gains do not rely on an unreported external computation budget.

\end{document}